\documentclass[conference]{IEEEtran}
\usepackage{times}

\usepackage[numbers]{natbib}
\usepackage{multicol}
\usepackage[bookmarks=true]{hyperref}
\usepackage{graphicx}
\usepackage{booktabs}    % Professional tables
\usepackage{multirow}    % Multi-row cells in tables
\usepackage{tabularx}    % Auto-width columns
\newcolumntype{Y}{>{\raggedright\arraybackslash}X}
\usepackage{amsmath,amssymb}  % Math symbols including \mathbb
\usepackage{placeins}
\usepackage{stfloats}
\usepackage{cuted}
\usepackage{capt-of}

\begin{document}

\title{RAPID: Reconfigurable, Adaptive Platform for Iterative Design}

\author{\authorblockN{Zi Yin\textsuperscript{*}\quad Fanhong Li\textsuperscript{*}\quad Shurui Zheng\quad Jia Liu\textsuperscript{\dag}}
\authorblockA{Tsinghua University\\[2pt]
{\normalfont\small \textsuperscript{*}Equal Contribution\quad \textsuperscript{\dag}Corresponding Author}}}

\twocolumn[{%
    \renewcommand\twocolumn[1][]{#1}%
    \maketitle
    \vspace{-4mm}
    \begin{center}
        \includegraphics[width=\textwidth]{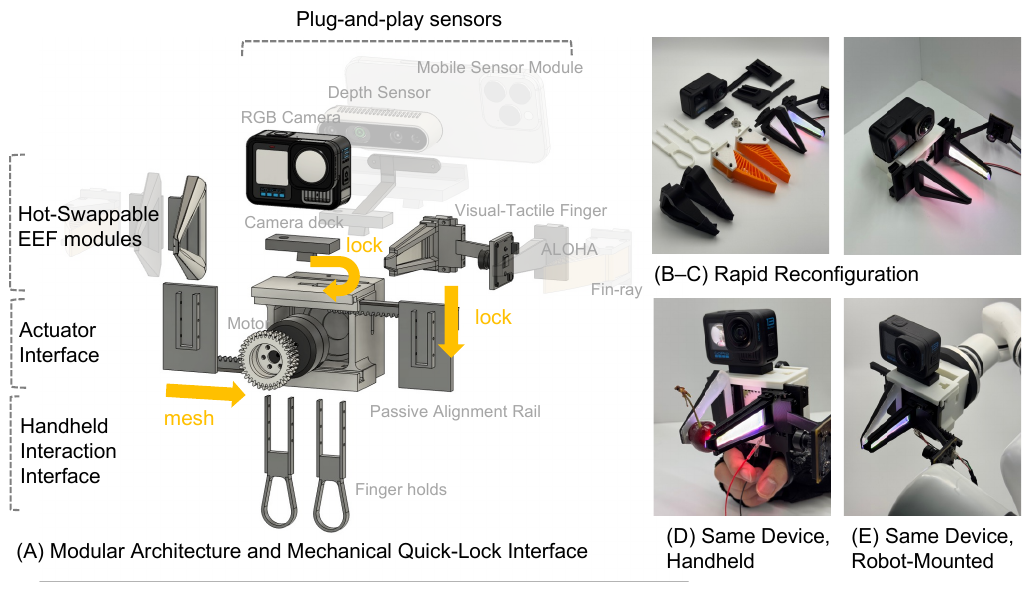}
        \captionof{figure}{RAPID system overview.
RAPID is a reconfigurable robotic platform enabling tool-free, plug-and-play integration of sensors, hot-swappable end-effectors, and modular actuation. The same modular device supports both handheld data collection and robot-mounted deployment through rapid physical reconfiguration.}
        \label{fig:overview}
    \end{center}
}]

\begin{abstract} Developing robotic manipulation policies is iterative and hypothesis-driven: researchers test tactile sensing, gripper geometries, and sensor placements through real-world data collection and training. Yet even minor end-effector changes often require mechanical refitting and system re-integration, slowing iteration. We present RAPID, a full-stack reconfigurable platform designed to reduce this friction. RAPID is built around a tool-free, modular hardware architecture that unifies handheld data collection and robot deployment, and a matching software stack that maintains real-time awareness of the underlying hardware configuration through a driver-level \emph{Physical Mask} derived from USB events. This modular hardware architecture reduces reconfiguration to seconds and makes systematic multi-modal ablation studies practical, allowing researchers to sweep diverse gripper and sensing configurations without repeated system bring-up. The Physical Mask exposes modality presence as an explicit runtime signal, enabling auto-configuration and graceful degradation under sensor hot-plug events, so policies can continue executing when sensors are physically added or removed. System-centric experiments show that RAPID reduces the setup time for multi-modal configurations by two orders of magnitude compared to traditional workflows and preserves policy execution under runtime sensor hot-unplug events.
The hardware designs, drivers, and software stack are open-sourced at \url{https://rapid-kit.github.io/}.
\end{abstract}

\IEEEpeerreviewmaketitle
\section{Introduction}

Learning robotic manipulation policies from demonstrations is increasingly an iterative, hypothesis-driven process: researchers test whether tactile sensing helps on fragile objects, whether wider fingers reduce failures, or whether alternative sensor placements improve observability in clutter~\cite{chi2023diffusion,zitkovich2023rt}. At the same time, the community has moved beyond visual-only imitation to multi-modal interfaces that combine vision, proprioception, and touch, often utilizing portable handheld grippers for in-the-wild data collection~\cite{chi2024universal,wu2023gello,fu2024mobile,zhao2023aloha}.

These handheld interfaces, from UMI and its variants~\cite{chi2024universal,zhaxizhuoma2025fastumi,xu2025dexumi,cheng2026tacumi,rayyan2025mvumi,ha2024umionlegs} to low-cost teleoperation systems~\cite{wu2023gello,fu2024mobile,zhao2023aloha}, have significantly broadened where and how we can collect demonstrations. However, as sensing and embodiment options proliferate, they also amplify a more mundane but critical bottleneck: every new gripper geometry, fingertip material, or sensor combination often means a new round of mechanical refitting and software integration before any data can be collected. In a typical study, characterizing the benefit of tactile sensing across different object geometries might require testing $N$ gripper designs against $M$ sensor configurations. Under a traditional workflow---typically involving screw-based end-effector swaps, manual configuration edits, and repeated restarts of the logging stack---systematically exploring this $N \times M$ space is prohibitively expensive.

We argue that to accelerate multi-modal manipulation research, we must transition from static interfaces to \emph{full-stack reconfigurable platforms}. In our experience, such a platform must solve two coupled problems: (i) the \emph{mechanical friction} of swapping components, so that changing end-effectors, fingertips, or sensor modules becomes a seconds-level operation rather than minutes of bespoke rework; and (ii) the \emph{modality observability gap}, namely the fact that the software stack has no principled way to know which sensing and actuation channels are physically present at a given moment, and therefore cannot safely adapt when sensors are added or removed.

We present RAPID (Reconfigurable, Adaptive Platform for Multi-Modal Iterative Design), a full-stack platform designed to reduce this iteration gap. RAPID is built around two pillars: (1) a tool-free, modular hardware architecture that unifies handheld data collection and robot-mounted deployment, allowing researchers to sweep across diverse gripper and sensing configurations without repeated system bring-up; and (2) a matching software stack that maintains real-time awareness of the underlying hardware configuration through a driver-level \emph{Physical Mask} derived from USB hot-plug events.

At the software level, the Physical Mask exposes modality presence as an explicit runtime signal: at any time step, the system knows which modalities (e.g., wrist camera, visuotactile fingertip, torque sensing) are physically online, based on USB connect and disconnect events. This hardware-grounded signal complements training-time modality dropout~\cite{neverova2015moddrop,liu2017sensor}. While dropout encourages policies to be robust to missing inputs at the representation level, the Physical Mask provides the system-level observability needed to auto-configure logging and to gracefully degrade policy execution when sensors are hot-plugged, allowing policies to continue executing when sensors are physically added or removed.

We evaluate RAPID through a system-centric lens, focusing on iteration overhead and runtime robustness rather than task-specific success rates. We quantify the time savings enabled by our tool-free design and auto-discovery pipeline in a representative multi-modal ablation scenario, and we probe runtime behavior by physically unplugging and re-plugging sensors during a tactile identification task. Our contributions are three-fold:

\begin{itemize}
\item \textbf{A full-stack reconfigurable manipulation platform} that unifies handheld data collection and robot-mounted deployment through a tool-free, modular hardware architecture and a unified I/O stack.

\item \textbf{A driver-level Physical Mask abstraction} that derives modality presence from USB events and exposes it as an explicit runtime signal, enabling auto-configuration and graceful degradation under sensor hot-plug events.

\item \textbf{A system-centric evaluation} demonstrating that RAPID reduces the setup time for multi-modal configurations by two orders of magnitude compared to traditional workflows, and preserves policy execution under runtime sensor hot-unplug scenarios.
\end{itemize}

\begin{figure*}[t]
    \centering
    \includegraphics[width=\textwidth]{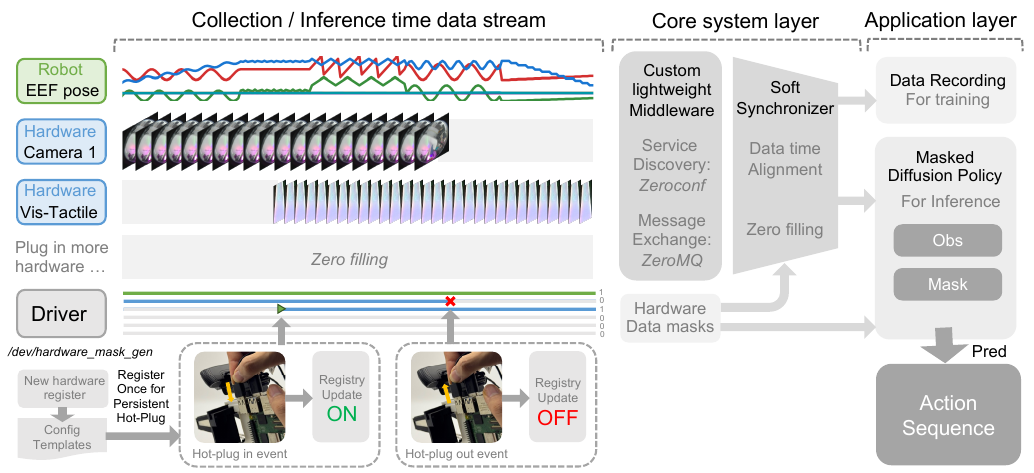}
\caption{Runtime architecture of the RAPID system.
The top panel shows asynchronous multimodal data streams during collection and inference, including robot end-effector state, vision, tactile sensing and optional other plug-in modalities.The bottom panel illustrates the event-driven pathway from hardware hot-plug events to the driver-generated Physical Mask, which is propagated through lightweight middleware to the application layer. The Physical Mask enables time alignment and zero filling of absent modalities, ensuring fixed-dimensional observations for data recording and mask-aware policy inference.}
    \label{fig:architecture}
\end{figure*}

\section{Related Work}
\label{sec:related_work}

\subsection{Portable Manipulation Interfaces and Hardware Fragmentation}

The Universal Manipulation Interface (UMI)~\cite{chi2024universal} established a paradigm shift in robot learning by enabling handheld, vision-based data collection without requiring the target robot during demonstration. This approach decouples demonstration capture from robot hardware, allowing ``in-the-wild'' data collection that has driven significant progress in imitation learning~\cite{chi2023diffusion,zitkovich2023rt}. 

The success of UMI has spurred a family of specialized variants, each addressing specific limitations. FastUMI~\cite{zhaxizhuoma2025fastumi} replaced offline visual SLAM with onboard Visual-Inertial Odometry using the Intel RealSense T265 to streamline tracking, though this introduced data format incompatibilities. DexUMI~\cite{xu2025dexumi} extended the concept to wearable hand exoskeletons, while TacUMI~\cite{cheng2026tacumi} integrated force-torque sensing and visuotactile fingertips, replacing the GoPro SLAM approach with HTC Vive external tracking to obtain \emph{interference-free} data. Other specialized variants include ExUMI~\cite{xu2025exumi} for proprioception via magnetic encoders, MV-UMI~\cite{rayyan2025mvumi} for multi-view capture, and UMI-on-Legs~\cite{ha2024umionlegs} for quadruped deployment. Parallel efforts like GELLO~\cite{wu2023gello} and Mobile ALOHA~\cite{fu2024mobile,zhao2023aloha} have similarly advanced low-cost teleoperation.

While each variant addresses real limitations, they exist largely as \emph{siloed designs}~\cite{bhirangi2024tactile}. The same UMI concept has spawned at least five fundamentally different tracking approaches (GoPro, T265, Vive, ARKit, Quest 3), each complicating direct data interoperability. Furthermore, adding new modalities often requires complete system redesign. In contrast, RAPID is designed not as a single fixed interface but as a \emph{full-stack reconfigurable platform} that aims to cover the continuous spectrum of configurations needed for systematic multi-modal ablation studies, while keeping the iteration friction of moving between them low.

\subsection{Multi-Modal Integration and Runtime Observability}

Integrating non-visual modalities into robot learning pipelines introduces significant system complexity. Tactile sensing improves manipulation in contact-rich tasks~\cite{lambeta2020digit,yuan2017gelsight,melnik2021using,shang2025forte}, but current architectures often treat sensor configurations as \emph{static assumptions}: the software expects a fixed set of sensors and fails ungracefully if one is missing~\cite{bednarek2020robustness}. This contributes to what we refer to as a \emph{modality observability gap}: the system has no principled way to know which modalities are physically present at a given time.

The foundational technique of modality dropout~\cite{neverova2015moddrop} and its adaptation to robotics~\cite{liu2017sensor} demonstrated that policies can learn robustness to sensor noise. Recent work has refined these techniques through disentangled representations (DisDP)~\cite{vanjani2025disdp}, variational information bottlenecks~\cite{du2022bayesian}, and inference-time composition (MCDP)~\cite{cao2025modality}. However, these works typically assume that sensor malfunctions are already detected by the system. Current fusion architectures, such as Visuo-Tactile Transformers~\cite{chen2022vtt}, FoAR~\cite{he2025foar}, and Visual-Geometry Diffusion Policy (VGDP)~\cite{tang2025visual}, provide \emph{algorithmic robustness} but do not make system-level observability of hardware a first-class concern. 
Similarly, runtime monitors like FIPER~\cite{romer2025fiper}, Sentinel~\cite{agia2024unpacking}, and uncertainty detectors~\cite{xu2025can} detect policy failures but largely assume valid data flow. The closest work on graceful degradation~\cite{sugiyama2025versatile} handles noisy sensors but not physical absence.

RAPID targets this system-level modality observability gap with a driver-level \emph{Physical Mask}. Unlike timeout heuristics that infer absence from missing data, the Physical Mask is updated directly from USB connect and disconnect events, exposing modality presence as an explicit runtime signal that can drive auto-configuration and graceful degradation when sensors are hot-plugged.

\subsection{Modular Design and Hardware Iteration Bottleneck}

The theoretical motivation for hardware-software co-design is well-established~\cite{sims2023evolving,chen2020hardware,chen2021codesign_focus}. However, co-design research is conducted almost exclusively in simulation because real-world iteration is prohibitively expensive~\cite{luck2020dataefficient,howard2019evolving,bhatia2021evogym}. The simulation-reality gap is often non-monotonic with respect to morphology~\cite{rosser2020sim2real,kriegman2019scalable}, meaning that predicting which hardware changes will improve real-world performance requires physical experimentation~\cite{aljalbout2025reality,koos2012transferability}.

Existing real-world efforts face significant integration friction. While some works have explored physical evolution~\cite{brodbeck2015morphological,nygaard2018real,cully2015robots}, manual assembly remains the bottleneck~\cite{moreno2021emerge,collins2021review}. Modular systems like the Yale OpenHand~\cite{ma2017yale} or Fable~\cite{pacheco2015fable} require significant reconfiguration time. Recent soft gripper systems~\cite{malik2025scalable} achieve fast swaps but are limited to specific actuators. RAPID adopts traditional joinery principles—specifically mortise-and-tenon joints~\cite{colle2025codesign,larsson2020tsugite,tosun2016design}—and pairs them with event-driven software discovery. Our goal is not autonomous self-reconfiguration, but minimizing the \emph{human overhead} of hardware iteration in a full-stack reconfigurable platform. By reducing per-configuration setup time from minutes to seconds, RAPID shifts the bottleneck of multi-modal ablation studies from engineering debugging back to scientific inquiry.

\begin{figure}[t]
    \centering
    \includegraphics[width=\columnwidth]{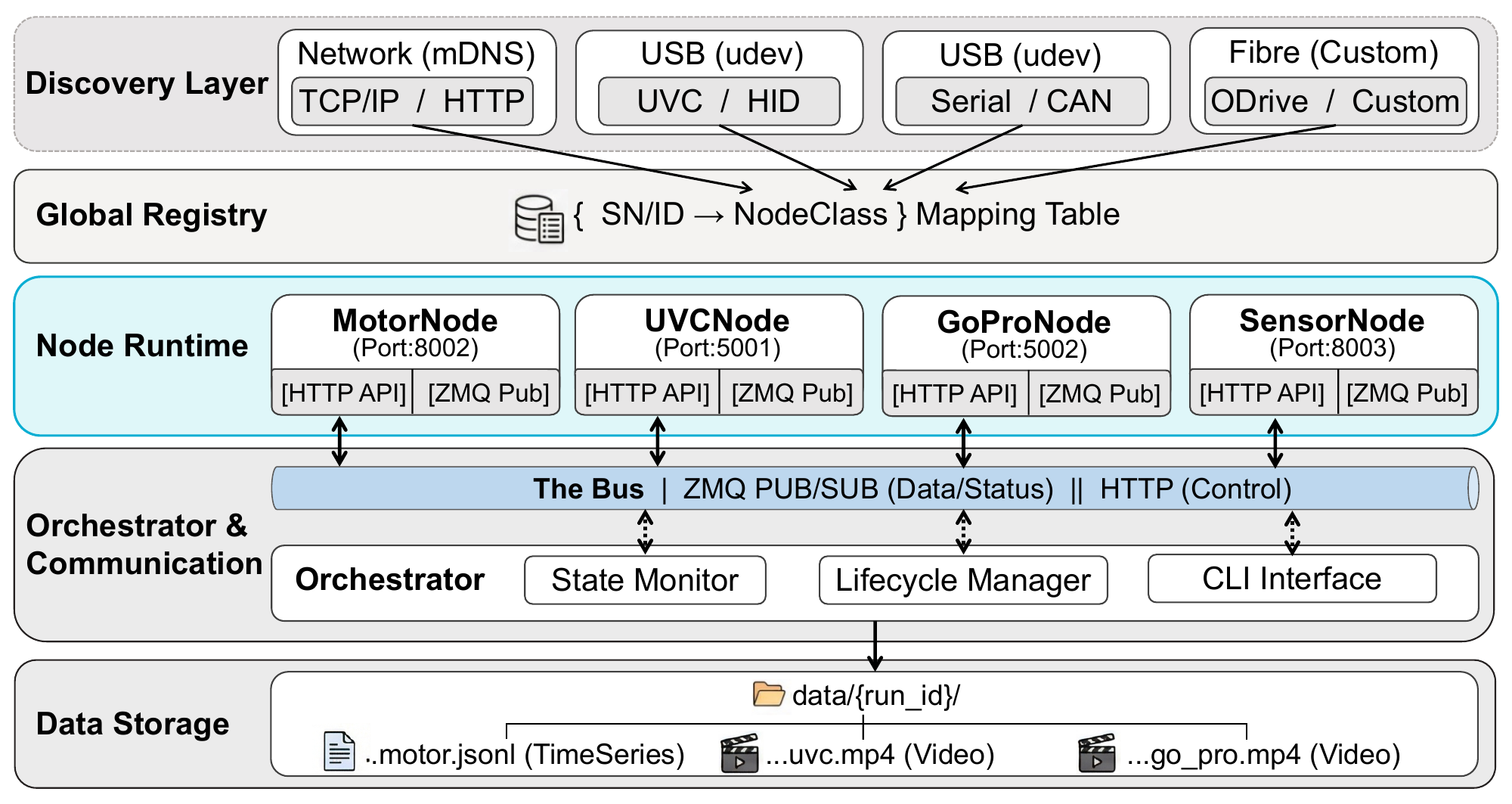}
    \caption{System implementation of RAPID. The discovery layer supports heterogeneous device interfaces (USB-CAN, USB-Serial, USB-LAN, and native USB) through a unified hub. The driver layer performs event-driven device registration and generates the Physical Mask via a virtual device file. The middleware layer publishes sensor streams and the mask over a lightweight publish-subscribe transport (ZeroMQ + Zeroconf). The synchronisation layer aligns multimodal observations within a configurable time window and zero-fills absent channels. The application layer consumes fixed-dimension observations in either collection mode (logging with per-frame mask) or inference mode (mask-aware policy execution).}
    \label{fig:implementation}
\end{figure}
\section{System Design}
\label{sec:system_design}

Building on the challenges identified in Section~\ref{sec:related_work}, RAPID is built around two pillars. The first is a \emph{tool-free, modular hardware architecture} (Section~\ref{subsec:hardware_layer}) that reduces end-effector and sensor swaps to seconds-level operations and unifies handheld data collection with robot-mounted deployment. The second is a \emph{driver-level Physical Mask} (Sections~\ref{subsec:driver_layer}--\ref{subsec:middleware_app}) that derives modality presence from USB hot-plug events, providing the system-level observability needed for auto-configuration and graceful degradation under runtime sensor changes. Fig.~\ref{fig:architecture} illustrates the resulting layered architecture.

\subsection{Hardware Layer}
\label{subsec:hardware_layer}

\begin{figure}[t]
    \centering
    \includegraphics[width=\columnwidth]{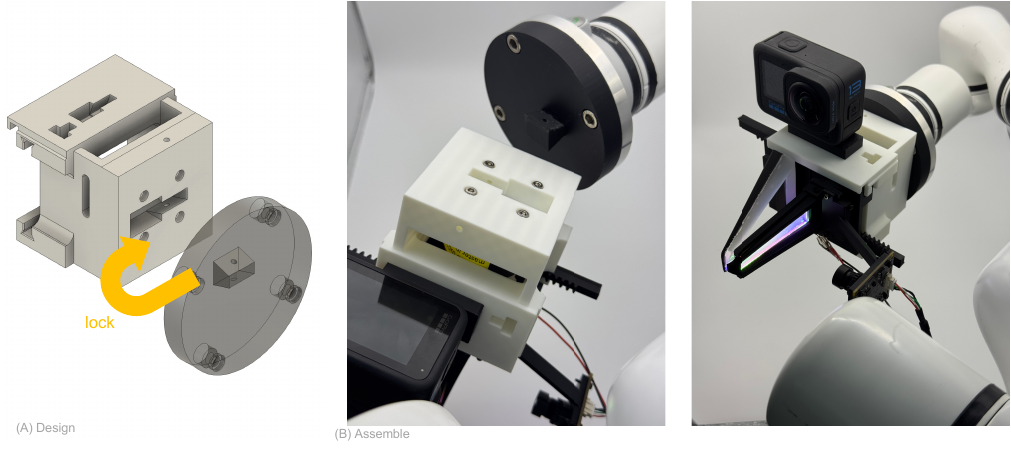}
	\caption{Hardware layer details. (A) Tool-free connector design. (B) Rapid deployment to robot arm as end-effector.}
    \label{fig:hardware}
\end{figure}

The hardware layer is designed to support frequent reconfiguration (Fig.~\ref{fig:hardware}).
\textbf{Base and standardized slots.} The base is a 3D-printed chassis that integrates motor mounting and cable routing; it remains fixed across configurations. The base exposes four standardized attachment slots: front (finger bodies), tip (fingertips), top (wrist camera mounts), and bottom (handheld or robot-mounting adapters). Only modules change; the base does not. This slot-based design also enables dual-use operation: the same base hardware supports both handheld data collection and robot-mounted deployment by swapping the bottom adapter.
\textbf{Mechanical attachment.} Modules connect via mortise-and-tenon joints, a geometry well-suited to FDM-printed PLA: it tolerates typical printing variation and survives repeated insertion cycles better than snap-fit alternatives. Similar joints appear in prior work, including the 3D-printed fingers of ALOHA~\cite{zhao2023aloha} and the GET tactile fingertip~\cite{burgess2025get}. For extended operation, a single set screw can be added for additional retention.
\textbf{Electrical interface.} We adopt USB as a unified interconnect layer. USB offers mature hot-plug support at the OS level and a wide ecosystem of converter boards, which simplifies replication. The system aggregates heterogeneous devices through a USB hub: USB2CAN (motors, robot arm), USB2Serial (ToF sensors), USB2LAN (LiDAR, GoPro), and native USB devices (RealSense, webcam, tactile sensors), as shown in Fig.~\ref{fig:implementation}. Devices with direct LAN connections are integrated through a parallel registration mechanism.
The output of this layer is the USB plug/unplug event: when a module is physically inserted or removed, the operating system generates an event that propagates to the driver layer. 
\subsection{Driver Layer}
\label{subsec:driver_layer}
The driver layer translates USB events into runtime state information for the software stack.
\textbf{Event detection and device registry.} A daemon process monitors USB plug/unplug events from the operating system. Upon each event, the daemon queries a registry of known modules; if the device matches a registered module, the corresponding data node is started or stopped. Stable device paths are established via udev symlinks (e.g., \texttt{/dev/rapid/tactile\_left}), using USB serial numbers when available or port topology as fallback. An orchestrator monitors node health via periodic heartbeats; failed nodes are automatically restarted with backoff.
\textbf{Single-point registration.} To add a new module type, users maintain a single configuration file specifying device identifiers and data publishing logic. A registration script validates the configuration and generates all derived artifacts (udev rules, descriptor files, node entries). This keeps the user-facing configuration minimal while ensuring system-wide consistency.

\textbf{Physical Mask Generator.} The central contribution of this layer is a virtual device that reads the current registry state and exposes the \textbf{Physical Mask} at 500 Hz via a device file (e.g., \texttt{/dev/hardware\_mask\_gen}). We define the Physical Mask as a vector-valued signal that maps each modality name to a boolean indicating its current physical presence, serialized as a JSON dictionary. Unlike data-timeout heuristics, this mask reflects hardware state directly: a modality is marked absent the moment its USB disconnect event fires, not after a timeout window (Fig.~\ref{fig:architecture}, bottom). This real-time, hardware-grounded signal is what enables downstream layers to distinguish transient packet loss from physical removal.
	The output of this layer is the Physical Mask device file, continuously updated to reflect which sensing and actuation channels are physically online.
	
\subsection{Middleware and Application Layer}
\label{subsec:middleware_app}

\textbf{Middleware.} We implement a lightweight publish-subscribe layer based on ZeroMQ and Zeroconf, chosen over ROS2~\cite{macenski2022ros2} for ease of customization and deployment in rapid prototyping. The Physical Mask device file from the driver layer is published as a topic at this layer, alongside sensor data streams.

\textbf{Synchronized Subscriber.} A synchronized subscriber aggregates multiple data topics and the mask topic, aligning them within a configurable time window (default 25\,ms, similar to ROS ApproximateSync), as illustrated in the three-layer data flow of Fig.~\ref{fig:architecture} (top). When the Physical Mask indicates a modality is offline, the subscriber performs zero-filling for that channel, ensuring downstream consumers always receive fixed-dimension input.

\textbf{Dual-mode operation.} The application layer supports two modes that share the same underlying infrastructure but run different programs:

\begin{itemize}
	\item \textit{Inference mode:} At each control step, the inference program fetches synchronized data with the current Physical Mask, then feeds both to a mask-aware multimodal Diffusion Policy~\cite{chi2023diffusion}. Absent modalities are zero-filled to maintain fixed input dimensions; the policy uses the mask to determine which channels carry valid observations. Because the mask is hardware-grounded, sensor hot-plug does not crash the pipeline.    
    \item \textit{Collection mode:} The collection program displays real-time modality status (registered vs.\ physically present), records synchronized data streams, and logs the Physical Mask per frame. Since the mask updates at 500\,Hz while data is recorded at 30--60\,Hz, the mask captures transient dropouts at finer granularity than the data itself. This enables systematic reuse of partially corrupted episodes and simplifies post-hoc data auditing.
\end{itemize}

Unlike training-time dropout masks derived through post-processing, the Physical Mask is recorded as ground-truth hardware state at capture time---information rarely available in existing datasets.

\section{Evaluation}
\label{sec:evaluation}

\textbf{Policy training.}
We use the standard CNN-based Diffusion Policy~\cite{chi2023diffusion} with two observation modalities: a wrist camera image and a visuotactile fingertip image. During training, each modality is independently dropped with probability $p{=}0.3$ (subject to the constraint that at least one modality remains), and the corresponding feature embedding is zeroed. At inference time, the Physical Mask replaces random dropout: when a sensor is physically absent, the same zeroing is applied to its feature embedding, exactly matching the training-time condition.

\subsection{Reconfiguration Efficiency}
\label{subsec:reconfig}

We evaluate the time savings enabled by RAPID's design choices in a representative ablation study scenario. Consider a researcher iterating through N=3 gripper materials (rigid, Finray, silicone) $\times$ M=3 sensor modalities (vision-only, vision+motor, vision+tactile), requiring 9 configurations total.

Table~\ref{tab:reconfig_time} compares the time cost under a traditional workflow versus RAPID. The traditional workflow is defined as RAPID with each design choice removed:
\begin{itemize}
    \item Without quick-release: screw-based end-effector swap
    \item Without auto-discovery: manual YAML configuration editing
    \item Without hot-swap: restart collection system via Ctrl-C and relaunch
    \item Without unified hardware: separate devices for handheld collection and robot execution, requiring redundant sensor deployment
\end{itemize}

In our prototype implementation and the configuration set used in this paper, RAPID reduces per-configuration setup time from approximately 480\,s to 5\,s. For the full N$\times$M=9 ablation, this translates to on the order of $100\times$ speedup, shifting the practical bottleneck from interface reconfiguration back to data collection itself.

\begin{table}[t]
\centering
\small
\caption{Reconfiguration time comparison: traditional workflow vs.\ RAPID for N$\times$M=9 configurations.}
\label{tab:reconfig_time}
\resizebox{\columnwidth}{!}{%
\begin{tabular}{@{}lccc@{}}
\toprule
\textbf{Design Choice} & \textbf{Traditional} & \textbf{$\Delta$ Saved} & \textbf{Ours} \\
\midrule
EEF swap & Screw (5$\pm$1 min) & $-$295 sec & Quick-release (5 sec) \\
Modality config & Manual YAML (2$\pm$0.5 min) & $-$120 sec & Auto-discovery (0 sec) \\
Collection system & Ctrl-C + restart (30$\pm$10 sec) & $-$30 sec & Hot-plug support (0 sec) \\
Sensor deployment & $\times$2 (handheld $\neq$ exec) & $-$50\% & $\times$1 (unified) \\
\midrule
\textbf{Per-config total} & $\sim$480 sec & $-$475 sec & $\sim$5 sec \\
\textbf{Full ablation (N$\times$M=9)} & $\sim$72 min & $-$71 min & $\sim$45 sec \\
\bottomrule
\end{tabular}}
\end{table}

\subsection{Inference with Runtime Modality Change}
\label{subsec:hotplug}

\begin{figure}[t]
    \centering
    \includegraphics[width=\columnwidth]{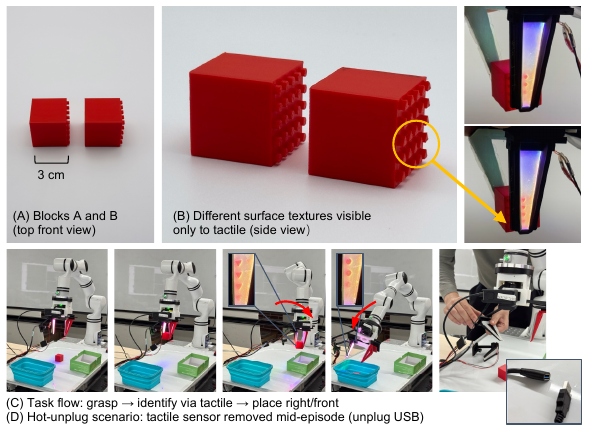}
    \caption{Experiment task for evaluating runtime modality change. Blocks A and B appear identical to the wrist camera but differ in surface texture (triangular vs.\ hexagonal) perceivable through tactile sensing. The policy must grasp, identify, and place the triangular block to the left and the hexagonal block to the front. We test system behavior when the tactile sensor is unplugged during execution.}
    \label{fig:experiment}
\end{figure}

We evaluate whether the Physical Mask mechanism enables graceful degradation during inference when sensors are physically added or removed (Fig.~\ref{fig:experiment}).
We emphasize that this experiment is not designed to benchmark manipulation performance against state-of-the-art policies, but to stress-test system behavior under runtime modality changes.

\textbf{Task.} The gripper is equipped with a wrist camera and a visuotactile fingertip. Two visually identical blocks (A and B) are presented; they differ only in surface texture on the side faces (one with a triangular pattern and the other with a hexagonal pattern), which is perceivable through tactile but not visual sensing. The policy must grasp each block, identify it via tactile feedback, and place the triangular block to the left and the hexagonal block to the front.

\textbf{Metric.} We report identification accuracy (correct placement / total trials) over 25 trials. We also report system status: Normal (completed), Degraded (completed with reduced performance), or Crash (pipeline failure).

\textbf{Conditions.} We compare RAPID (with Physical Mask) against a baseline that uses the same hardware and policy but replaces the Physical Mask layer with a static configuration file specifying which sensors are expected. In the baseline, changing the sensor set requires manually editing this file and restarting the pipeline before execution. We test four conditions:
\begin{itemize}
    \item \textit{Full}: wrist camera + tactile online throughout.
    \item \textit{No-tactile}: tactile absent from the start. For the baseline, the tactile entry is manually removed from the configuration file before launch.
    \item \textit{Hot-unplug}: tactile physically removed mid-episode.
    \item \textit{Hot-replug}: tactile removed and then reinserted mid-episode.
\end{itemize}

\textbf{Results.} Table~\ref{tab:hotplug} reports results per condition.

\begin{table}[t]
\centering
\small
\caption{Inference under runtime modality changes. Score: identification accuracy over 25 trials. Status: Normal / Degraded / Crash.}
\label{tab:hotplug}
\begin{tabular}{@{}lcccc@{}}
\toprule
\multirow{2}{*}{\textbf{Condition}} & \multicolumn{2}{c}{\textbf{RAPID}} & \multicolumn{2}{c}{\textbf{Static Config}} \\
\cmidrule(lr){2-3} \cmidrule(lr){4-5}
& Score & Status & Score & Status \\
\midrule
Full        & 0.92 & Normal   & 0.92 & Normal   \\
No-tactile  & 0.52 & Degraded & 0.52 & Degraded \\
Hot-unplug  & 0.56 & Degraded & —    & Crash    \\
Hot-replug  & 0.84 & Normal   & —    & Crash    \\
\bottomrule
\end{tabular}
\end{table}
% ------------------------------
\section{Discussion}
\label{sec:discussion}

We presented RAPID to address the high integration friction that currently hinders systematic multi-modal ablation studies. By decoupling component exchange from system reconfiguration, RAPID shifts the bottleneck of the research loop from engineering debugging back to scientific inquiry.

\subsection{Trade-offs of Tool-Free Modularity}
A primary concern with reconfigurable hardware is the potential loss of mechanical rigidity compared to screw-based assemblies. Our use of mortise-and-tenon joints prioritizes exchange speed and 3D printing tolerance over absolute stiffness. While sufficient for the contact forces typical of handheld data collection and pick-and-place tasks, this design may introduce deflection under high-load manipulation or precision assembly tasks. For such regimes, the system supports adding a single retention screw to lock the joint, effectively trading reconfiguration speed for mechanical rigidity. This hybrid approach allows users to use the tool-free mode for rapid prototyping and the locked mode for final large-scale data collection. A more systematic characterization of this stiffness--reconfigurability trade-off, across different materials and loading regimes, is left for future work.

\subsection{Physical Mask as a Data Curation Tool}
Beyond preventing runtime crashes, the Physical Mask has broader implications for data efficiency. In traditional pipelines, episodes with missing sensor data are often discarded entirely to maintain tensor shape consistency. RAPID enables the collection and utilization of heterogeneous datasets where episodes with varying sensor availability coexist. By logging the mask alongside the trajectory, researchers can train policies that explicitly learn robustness to sensor failure or leverage expensive tactile sensors only for a subset of the data while using cheaper visual-only setups for the bulk. This capability transforms hardware failure from a fatal error into a manageable data attribute. In this sense, the Physical Mask is not only a runtime safety mechanism but also a dataset-level abstraction that supports more realistic system-centric evaluation of multimodal policies.

\subsection{Limitations}
We acknowledge several limitations in both our system implementation and experimental scope.
% [Technical Limitations]
Technically, relying on USB hubs to aggregate heterogeneous sensors introduces bandwidth constraints when operating multiple high-resolution cameras simultaneously. Furthermore, while our software synchronization is effective for standard policy learning, it lacks the microsecond-level precision of hardware-triggered synchronization required for high-speed dynamic manipulation.
% [Experimental Scope Limitations - Moved from Section 4.3]
Experimentally, our evaluation was conducted in a single laboratory setting. While the system aims to democratize data collection, large-scale cross-institution validation is required to fully characterize reproducibility. Additionally, we do not claim state-of-the-art manipulation performance; our experiments were designed to stress-test system stability and iteration friction rather than to benchmark policy success rates against highly-tuned baselines. Moreover, while RAPID automates device discovery, the geometric alignment of newly inserted cameras still requires manual extrinsic calibration, which remains a per-reconfiguration bottleneck. Finally, the long-term durability of the PLA printed interfaces under thousands of insertion cycles remains to be quantified.

\section{Conclusion}
\label{sec:conclusion}

In this work, we introduced RAPID, a reconfigurable platform designed to accelerate multi-modal manipulation research. By mirroring mechanical modularity with software modality observability, we addressed the integration overhead that plagues iterative hardware design. Our system-centric evaluation demonstrated that RAPID reduces the time required for hardware reconfiguration by two orders of magnitude compared to traditional workflows and enables robust inference even when sensors are physically removed.

These results suggest that treating data collection interfaces as adaptive platforms rather than static devices is a viable path toward more rigorous real-world hardware--software co-design. We hope that by lowering the barrier to hardware iteration, RAPID will encourage the community to move beyond fixed benchmarks and explore the vast design space of robot morphology and sensing. We have open-sourced the complete hardware design files and software stack to facilitate this transition.

In future work, we aim to address the remaining bottleneck of extrinsic calibration. While RAPID automates device discovery, the geometric alignment of new cameras currently requires manual recalibration. We plan to integrate marker-based self-calibration routines that trigger automatically upon module insertion, further closing the loop on fully autonomous reconfiguration.

\section*{Acknowledgments}
We thank Yun Gui and Longsheng Jiang for their valuable discussions, feedback, and support throughout this project.

\bibliographystyle{plainnat}
\bibliography{references}

\clearpage
%% ==================== APPENDIX ====================
\appendix

\section{System Implementation Details}
\label{app:system_details}

\subsection{Physical Mask Data Format}
\label{app:mask_format}

The main text describes the Physical Mask as a ``JSON dictionary.'' For clarity, we provide detailed specifications of both the actual runtime format and the debug format:

\textbf{Binary Format (Runtime).} The production system publishes the Physical Mask at 500\,Hz via shared memory (\texttt{/dev/shm/rapid\_hardware\_mask}) for minimal latency. The binary layout is cache-line aligned (32 bytes):

\begin{table}[h]
\centering
\small
\caption{Physical Mask binary format (32 bytes, cache-line aligned).}
\label{tab:mask_binary}
\begin{tabularx}{\columnwidth}{@{}ccl>{\raggedright\arraybackslash}X@{}}
\toprule
\textbf{Offset} & \textbf{Size} & \textbf{Field} & \textbf{Description} \\
\midrule
0  & 4 & magic         & 0x52415044 (``RAPD'') \\
4  & 1 & version       & Protocol version (1) \\
5  & 1 & device\_count & Number of registered devices \\
6  & 2 & \_padding     & Alignment padding \\
8  & 8 & mask          & Online status bitmask (u64) \\
16 & 8 & timestamp\_ns & Nanosecond timestamp \\
24 & 8 & sequence      & Monotonic sequence number \\
\bottomrule
\end{tabularx}
\end{table}

\textbf{JSON Format (Debug).} For debugging and visualization, a separate process outputs a human-readable JSON file at 1\,Hz to \texttt{/dev/shm/rapid\_hardware\_mask.json}:

{\footnotesize
\begin{verbatim}
{
  "timestamp": "2026-02-05T10:30:00Z",
  "device_count": 3,
  "online_count": 2,
  "mask": "0x05",
  "mask_binary": "00000101",
  "sequence": 123456,
  "devices": [
    {"name": "cam_wrist",
     "bit": 0, "online": true},
    {"name": "tac_left",
     "bit": 1, "online": false},
    {"name": "motor_grip",
     "bit": 2, "online": true}
  ]
}
\end{verbatim}
}

\subsection{Device Discovery Pipeline}
\label{app:discovery}

The discovery pipeline employs a two-tier matching strategy:

\begin{enumerate}
    \item \textbf{Exact match:} VID + PID + Serial number uniquely identifies a specific device instance (e.g., a particular tactile sensor).
    \item \textbf{Model match:} VID + PID identifies the device type when serial numbers are unavailable or for interchangeable devices.
\end{enumerate}

Device registration is specified via TOML configuration files:

{\footnotesize
\begin{verbatim}
[device.tactile_left]
vid = "0x1234"
pid = "0x5678"
serial = "TACL001"
node = "tactile_publisher"
topic = "/rapid/tactile/left"
\end{verbatim}
}

The registration script automatically generates udev rules (e.g., \texttt{/etc/udev/rules.d/99-rapid.rules}) to create stable symlinks such as \texttt{/dev/rapid/tactile\_left}.

\subsection{System Latency Analysis}
\label{app:latency}

Table~\ref{tab:latency} summarizes the end-to-end latency from hardware event to software response.

\begin{table}[h]
\centering
\small
\caption{System latency breakdown for device hot-plug events.}
\label{tab:latency}
\begin{tabularx}{\columnwidth}{@{}Xc@{}}
\toprule
\textbf{Stage} & \textbf{Latency} \\
\midrule
USB insertion $\rightarrow$ OS udev event & 50--100\,ms \\
udev event $\rightarrow$ process startup & 100--200\,ms \\
Process startup $\rightarrow$ data stream ready & 800\,ms--1.5\,s \\
\midrule
\textbf{End-to-end (USB $\rightarrow$ visualization)} & \textbf{1.0--1.8\,s} \\
\bottomrule
\end{tabularx}
\end{table}

The ``process startup to data stream ready'' latency is device-dependent: USB cameras typically initialize within 800\,ms, while network-connected devices (e.g., GoPro via USB-LAN adapter) may require up to 1.5\,s. The Physical Mask itself is updated at 500\,Hz (2\,ms interval), ensuring that modality presence information is available with minimal delay after the OS-level event.

\subsection{Driver Implementation}
\label{app:driver}

The device management system is implemented as a Rust daemon (\texttt{rapid\_driver}) that monitors USB events, manages sensor processes, and publishes the Physical Mask to shared memory.

\begin{figure}[h]
    \centering
    \includegraphics[width=0.9\columnwidth]{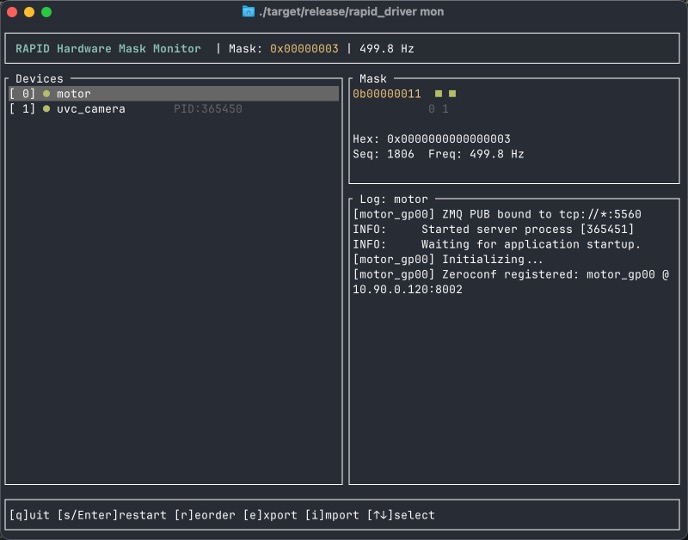}
    \caption{Terminal interface showing Physical Mask monitor with real-time device status, bitmask visualization, and process logs. Each device displays connection state (green: online, yellow: USB connected but process starting, red: offline) and assigned bit position.}
    \label{fig:hard_mask}
\end{figure}

\textbf{Shared Memory Mask Publishing.}
The Physical Mask is published at 500\,Hz to \texttt{/dev/shm/rapid\_hardware\_mask} as a 32-byte, cache-line aligned binary structure (Table~\ref{tab:mask_binary}). The structure uses a magic number (\texttt{0x52415044}, ``RAPD'') for validation, a monotonic sequence number for detecting stale reads, and nanosecond timestamps for synchronization. The 64-bit mask field supports up to 64 independently tracked devices. A separate JSON file is published at 1\,Hz for debugging and visualization purposes (Figure~\ref{fig:hard_mask}).

\textbf{Hot-Plug Event Handling.}
Device lifecycle management is driven by \texttt{udev} events from the Linux kernel. When a USB device is inserted, the daemon extracts its identity (VID, PID, serial number) and performs two-pass matching against the registry:
\begin{enumerate}
    \item \textbf{Exact match:} VID + PID + serial number uniquely identifies a specific device instance.
    \item \textbf{Model match:} VID + PID alone identifies interchangeable devices when serial numbers are unavailable.
\end{enumerate}
Upon successful match, the daemon spawns the configured \texttt{on\_attach} command (typically a ZeroMQ publisher node) as a child process with its own process group. A 2-second cooldown prevents rapid reconnection loops from USB electrical bounce.

\textbf{Graceful Process Termination.}
When a device is unplugged, the daemon terminates the associated process using a graceful shutdown sequence: \texttt{SIGTERM} is sent to the process group, followed by a 5-second grace period for cleanup. If the process does not exit within this window, \texttt{SIGKILL} is issued to force termination. An optional \texttt{on\_detach} command can be specified for additional cleanup (e.g., releasing network resources). Crashed processes are automatically restarted with exponential backoff (up to 5 attempts within 60 seconds) if the USB device remains connected.

\begin{figure}[h]
    \centering
    \includegraphics[width=0.9\columnwidth]{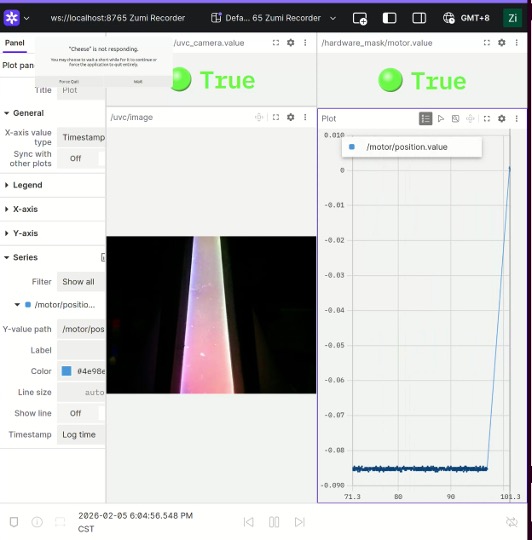}
    \caption{Foxglove Studio visualization of MCAP-recorded session showing synchronized multi-modal data: wrist camera view, tactile sensor images, and robot joint states. The timeline enables scrubbing through recorded episodes for analysis and debugging.}
    \label{fig:foxglove}
\end{figure}

\textbf{Data Recording and Visualization.}
Sensor data streams are recorded in MCAP format, an open-source container format designed for robotics data. MCAP provides efficient columnar storage, random-access playback, and native support for ROS-style messages. The recorded data can be directly loaded into Foxglove Studio for multi-modal visualization (Figure~\ref{fig:foxglove}), enabling synchronized playback of camera feeds, tactile images, and robot state. This recording pipeline captures the Physical Mask alongside sensor data, allowing offline analysis of how modality availability affects policy behavior.

%% ==================== POLICY TRAINING ====================
\section{Policy Training Details}
\label{app:training}

\subsection{Dataset Statistics}
\label{app:dataset}

\begin{table}[h]
\centering
\small
\caption{Dataset statistics for the tactile identification task.}
\label{tab:dataset}
\begin{tabularx}{\columnwidth}{@{}Xc@{}}
\toprule
\textbf{Parameter} & \textbf{Value} \\
\midrule
Total episodes & 72 \\
Episodes per configuration & 36 \\
Frames per episode & $\sim$100--170 \\
\bottomrule
\end{tabularx}
\end{table}

The dataset comprises demonstrations collected with two block types (triangular and hexagonal texture), with 36 episodes per type. Episodes vary in length depending on grasp timing and placement trajectory.

\subsection{Diffusion Policy Configuration}
\label{app:diffusion_config}

\begin{table}[h]
\centering
\small
\caption{Diffusion Policy hyperparameters.}
\label{tab:diffusion_params}
\begin{tabularx}{\columnwidth}{@{}Xc@{}}
\toprule
\textbf{Parameter} & \textbf{Value} \\
\midrule
Architecture & CNN-based Diffusion Policy \\
Observation horizon & 2 \\
Action horizon & 8 \\
Prediction horizon & 16 \\
Training epochs & 200 \\
Modality dropout rate & 0.3 \\
\bottomrule
\end{tabularx}
\end{table}

\subsection{Ablation Study: Modality Configurations}
\label{app:ablation}

Table~\ref{tab:ablation} presents ablation experiments investigating the contribution of different input modalities.

\begin{table}[h]
\centering
\footnotesize
\caption{Ablation study on modality configurations.}
\label{tab:ablation}
\begin{tabularx}{\columnwidth}{@{}c>{\raggedright\arraybackslash}Xccl@{}}
\toprule
\textbf{\#} & \textbf{Config} & \textbf{Data} & \textbf{Drop} & \textbf{Result} \\
\midrule
1 & Vision only & mix & -- & Biased (0.50) \\
2 & Vision + mask & mix & 0.3 & Biased (0.50) \\
3 & Vis + mask + diff & mix & 0.3 & \textbf{Success (1.00)} \\
4 & Vision + diff & big & -- & Biased (0.50) \\
5 & Vision + diff & mix & -- & \textbf{Success (1.00)} \\
\bottomrule
\end{tabularx}
\\[0.5em]
\raggedright\footnotesize
Data: mix = mixture, big = bigcube. All diff image experiments use hand\_view. ``Biased'' = policy places all objects to the same side; ``Success'' = correct texture-based placement.
\end{table}

\textbf{Key findings:}
\begin{itemize}
    \item \textbf{Experiments 1--2:} Vision alone cannot discriminate objects that appear identical but differ in tactile texture.
    \item \textbf{Experiments 3, 5:} The diff image effectively captures tactile-induced visual changes, enabling successful texture identification.
    \item \textbf{Experiment 4 vs.\ 5:} Dataset diversity (mixture vs.\ single configuration) affects policy generalization; training on a single cube size leads to biased behavior.
\end{itemize}

\subsection{Diff Image Computation}
\label{app:diff_image}

The diff image captures visual changes induced by tactile contact:
\begin{equation}
    \text{diff} = \frac{\text{current\_frame} - \text{reference\_frame}}{2} + 0.5
\end{equation}
where \texttt{reference\_frame} is the first frame of the episode (before contact). The output is normalized to $[0, 1]$, with 0.5 representing no change. This encoding preserves both positive and negative intensity changes while maintaining a consistent input range for the policy network.

%% ==================== TACTILE CLASSIFICATION ====================
\section{Tactile Classification Auxiliary Experiment}
\label{app:classification}

To verify that tactile images contain discriminative texture information, we trained a standalone classifier on tactile frames.

\begin{table}[h]
\centering
\small
\caption{Tactile texture classification results.}
\label{tab:classification}
\begin{tabularx}{\columnwidth}{@{}Xc@{}}
\toprule
\textbf{Metric} & \textbf{Value} \\
\midrule
Architecture & ResNet-18 (ImageNet pretrained) \\
Dataset split & Episode-level (80/10/10) \\
Test accuracy & \textbf{99.73\%} \\
Precision (triangular / hexagonal) & 1.000 / 0.994 \\
Recall (triangular / hexagonal) & 0.995 / 1.000 \\
\bottomrule
\end{tabularx}
\end{table}

\textbf{Confusion Matrix:}
\begin{table}[h]
\centering
\small
\begin{tabularx}{\columnwidth}{@{}Xcc@{}}
\toprule
& \textbf{Pred: Triangular} & \textbf{Pred: Hexagonal} \\
\midrule
\textbf{True: Triangular} & 586 & 3 \\
\textbf{True: Hexagonal} & 0 & 536 \\
\bottomrule
\end{tabularx}
\end{table}

The near-perfect classification accuracy confirms that tactile images capture sufficient information to discriminate the two textures. This validates that the policy's inability to distinguish textures in vision-only conditions (Table~\ref{tab:ablation}, Experiments 1--2) stems from the visual similarity of the blocks rather than insufficient tactile information in the diff images.

%% ==================== MODULE CATALOG ====================
\FloatBarrier
\section{Module Catalog}
\label{app:modules}

Table~\ref{tab:modules} lists the hardware modules used in this work. Sensor-bearing modules provide YAML descriptors for automatic discovery; mechanical-only modules require no software configuration.

\begin{center}
\footnotesize
\captionof{table}{RAPID module catalog.}
\label{tab:modules}
\begin{tabularx}{\columnwidth}{@{}Xlll@{}}
\toprule
\textbf{Module} & \textbf{Slot} & \textbf{Sensors} & \textbf{Bring-up} \\
\midrule
Finger (parallel-jaw) & front & -- & -- \\
Finger (pinch) & front & -- & -- \\
Rigid tip (PLA) & tip & -- & -- \\
Soft tip (TPU) & tip & -- & -- \\
Tactile tip (GET) & tip & cam+LED & USB \\
Wrist mount (std) & top & GoPro & Net \\
Wrist mount (angled) & top & GoPro & Net \\
Handheld adapter & bottom & -- & -- \\
Robot adapter & bottom & -- & -- \\
\bottomrule
\end{tabularx}
\\[0.5em]
\raggedright\footnotesize
Slots: front (finger bodies), tip (fingertips), top (wrist mounts), bottom (adapters). Sensor-bearing modules provide YAML descriptors for automatic discovery; mechanical-only modules require no software configuration.
\end{center}

\end{document}